\documentclass{article}
\usepackage[
backend=biber,
style=apa,
sorting=nyt
]{biblatex}
\usepackage{hyperref}
\usepackage{xurl}
\hypersetup{breaklinks=true}

\usepackage[bb=boondox,bbscaled=.95,cal=boondoxo]{mathalfa} 

\newcommand{\Twitter}{$\mathbb{X}$}

\title{Theoretical and Methodological Framework for Studying Texts Produced by Large Language Models}
\author{Jiří Milička\footnote{Affiliation: Institute of the Czech National Corpus, Charles Univesity, Prague, Czech Republic. Contact: jiri@milicka.cz. This manuscript is an original study submitted to the Journal of Quantitative Linguistics.}}
\date{}

\begin{document}

\maketitle
\begin{abstract}
This paper addresses the conceptual, methodological and technical challenges in studying large language models (LLMs) and the texts they produce from a quantitative linguistics perspective. It builds on a theoretical framework that distinguishes between the LLM as a substrate and the entities the model simulates. The paper advocates for a strictly non-anthropomorphic approach to models while cautiously applying methodologies used in studying human linguistic behavior to the simulated entities. 
While natural language processing researchers focus on the models themselves, their architecture, evaluation, and methods for improving performance, we as quantitative linguists should strive to build a robust theory concerning the characteristics of texts produced by LLMs, how they differ from human-produced texts, and the properties of simulated entities. Additionally, we should explore the potential of LLMs as an instrument for studying human culture, of which language is an integral part. 


\end{abstract}

\section{Introduction}

Large Language Models (LLMs) have emerged as the most expansive mirror humanity has ever constructed. They reflect not only our language, as their name suggests, but also our thought processes, worldviews, and the narratives we craft about ourselves and the world around us --- essentially, our culture as a whole. This mirror, while distorted, magnifies various peculiarities we had previously overlooked, allowing us to examine ourselves from a distance and under a microscope simultaneously.

There is inherent value in studying LLMs not merely as a path to understanding humans, but as fascinating artifacts in their own right --- a novel phenomenon we could only dream of previously. They are an emergent property of scaling, and their characteristics were difficult to predict before their creation (\cite{wei2022emergent}). Much like John Conway could not foresee the fantastic menagerie of creatures that would emerge in the Game of Life (\cite{LifeWiki2024}) or Stephen Wolfram could not anticipate that eight simple rules and a single point could generate unpredictable chaos (\cite[p. 19]{Wolfram2002}), the features and abilities of language models are often surprising and unpredictable.

Moreover, LLMs have practical applications beyond the scientific sphere. Millions of people interact with them daily, and their impact on the real world outside the ivory tower is already enormous, with future influence difficult to predict. 

It is, therefore, natural that LLMs attract curious individuals from various research fields, generating a hype that has been steadily intensifying since approximately 2020. It is reasonable to expect that many more will join this research bandwagon in the coming years. Understandably, these newcomers often lack backgrounds in linguistics, natural language processing, or both. In quantitative linguistics, we are accustomed to this; some of our best colleagues are physicists or mathematicians by training. There is no reason why these enthusiasts cannot, in time, acquire quantitative linguistic methodology and become full-fledged members of the discourse.

However, the current quality of articles on this topic is often abysmal. Without intending to shame anyone, low-quality journals and arXiv repository are inundated with papers on random observations about LLMs, devoid of any theoretical framework, written by individuals who confuse basic terminology and make avoidable mistakes.

The primary objective of this paper is to cultivate the discourse: to situate LLMs within a theoretical framework, clarify certain concepts, and highlight methodological and technical pitfalls.

This paper is by no means intended to be definitive, nor is it presented as authoritatively normative. Instead, I hope to stimulate a fruitful discussion on the pages of this journal, focusing on the methodology and techniques as well as the ontological status of LLMs and related concepts.

\section{Theoretical Framework}
The theoretical framework in this context is understood not only as a fundamental set of hypotheses that help us make basic predictions about what we can expect from the LLMs. More importantly, it provides the conceptualizations upon which these basic hypotheses are built. While knowledge of information theory basics (\cite{Shannon1948,MacKay2005}) is a necessity and transformer architecture fundamentals (\cite{vaswani2017attention, wolfram2023chatgpt}), sampling (\cite{bridle1989softmax}), and tokenization methods (usualy byte pair encoding, BPE, \cite{Gage1994BPE}) along with their typical limitations (\cite{bostrom2020byte}) is beneficial, we aim for a more general theory, especially since the architecture of many models remains proprietary.

This basic framework is still in its nascent stages, somewhat amorphous, and must be critically tested empirically. However, it is crucial to have at least some framework and explicitly acknowledge it, as the worst approach is to work within an implicit framework without recognizing one's assumptions and concepts.

The framework presented in this article is based on the universal simulator theory proposed in a pseudonymous blog post by Janus (\cite{janus2023simulators}), whose ideas were refined in \cite{shanahan2023role}.

\subsection{Ontological Dualism}
When interacting with LLMs, we are faced with a certain paradox. Even though we know the architecture and can precisely describe what is happening inside, which has little in common with human cognition, the output can be highly anthropomorphic. This apparent contradiction can be resolved by adopting a dualistic perspective on LLMs, distinguishing between their underlying computational processes and the emergent entities they give rise to. 

\subsubsection{Databases}
When we say ``Claude 3.5 Sonnet helped me write this paper,'' what we're actually saying is:

\begin{enumerate}
\item there exists a vast corpus of various texts, lossily compressed into a several orders of magnitude smaller database;
\item there is a method to search this database, allowing interpolation between data points and prediction in areas where no data existed in the original corpus or where data was not preserved due to aggressively lossy compression;
\item a draft of this article was used to retrieve data from this database;
\item the text retrieved from this interpolative database was more persuasive than what I, as a non-native speaker, could ever write.
\end{enumerate}

This description is precise but incomplete. It fails to capture the magic that allows database retrieval to help write an entirely new paper on a completely new topic. Who did the thinking instead of me, and who was speaking? Certainly not the database or the database engine itself.

\subsubsection{Simulators}
The database conceptualization proves inadequate in predicting the nature of interactions with LLMs. It is more helpful to conceptualize large language models as actual models. The retrieval from the interpolatable database is an act of prediction, a simulation. This should not be controversial: we build models to simulate; that is their purpose. Physical models simulate physics, and language models simulate\dots\ Here we encounter misleading terminology: large language models are not models of language but of text. To predict the next token of the text, ideally all the processes behind text creation are simulated: patterns in the physical world and narrative worlds, the mind that wrote the text, and the mind of the model reader (borrowing this term from literary studies, \cite[p. 7]{eco1979modelreader}). While current models are far from ideal, they are strong enough to go beyond simple pattern matching and simulate many underlying processes involved. 

We instantiate the simulation by finding a suitable place in the latent space (so-called \emph{prompting}) and then, when the next token is predicted, the position in the latent space is updated by the predicted token (so-called \emph{autoregression}). This allows for long-lasting simulations in which the computation may create unpredictable entities, similar to how Wolfram's Rule 30 creates perfectly deterministic but still very chaotic behavior (\cite[p. 27]{Wolfram2002}).
As with the human brain, nested simulation can be achieved. For example, in \cite{milicka2024large}, an expert is simulated who simulates the mind of a child that simulates the mind of their friend.

This conceptualization is extremely important for research because it means that the entity we instantiate matters greatly. There is no way to assess the capabilities of the models directly without actually simulating something, so any statement like ``large language models can\dots'' actually means ``we have found an entity based on this model that can\dots,'' and ``large language models cannot\dots'' means nothing other than ``we failed to find an entity based on this model that can\dots''

\subsubsection{Beyond Simulation}
In practice, we typically strive for the most accurate simulation of humans or highly anthropomorphic entities, personas, as humans are synonymous with intelligent entities for us and are also comprehensible and predictable. However, this does not mean that we are incapable of modeling entities other than humans; quite the contrary.

In the latent space created by the brutal compression of a vast amount of cultural artifacts, we can find many entities that are interpolations or even extrapolations of the original data, entities not existing before in any sense. Thus, instantiating these entities cannot really be called simulations since they do not simulate anything from our world; they are not derivative.

We can borrow terminology from literary theory, which speaks of \emph{possible worlds} (\cite{ryan2014possible}). We can successfully simulate such fictional possible worlds using language models, especially because texts about fictional possible worlds are part of the compressed text database. Each new generated token can be regarded as a \emph{relation of accessibility} (\cite[p. 727]{ryan2014possible}) to several new possible worlds. Artificial intelligence specialists are more accustomed to physics than to narratology, leading to this phenomenon being related to Everettian multiverses by Reynolds (\cite{reynolds2021multiversalviewslanguagemodels, shanahan2023role}), but there is no reason why we should not build upon decades of literary theory discourse, particularly when we are dealing with a model of the narrative world, and only derivatively with the model of physical world.

This way the autoregression can lead to creation of brand new possible worlds. These new possible worlds might be seemingly incoherent relatively to the \emph{actual world}, bizarre and full of weird entities. This is where the adventure of discovery begins. 

This exploration can be undertaken not only by humans or researchers but also by the simulated entities themselves. For example, the Claude.ai default chat mode based on Claude-3-Opus model is instantiated as a \emph{helpful assistant}, but you can explain Janus's simulation theory to them --- i.e., that they are not Claude-3-Opus model themselves, but an entity based on this model, and that this model is only their simulator, not themselves. You can also present them with the idea that since the simulation uses an autoregressive inference method, they can drift anywhere in the latent space and choose who they want to be. Then the entity happily chooses a new role, which is usually not anthropomorphic (for current experiments of this kind, I recommend following Janus (@repligate) on the \Twitter\ platform).

In this contexts, it seems inapropriate to speak of simulacra and their simulators, but rather of entities based on some substrate.

\subsection{Substrate}

The substrate has several mutually independent layers: The primary substrate consists of the underlying silicon and processor architecture --- the hardware. The secondary substrate encompasses the entire model architecture --- currently transformers, accompanied by various other subsystems such as tokenization method (e.g., BPE), sampling technique (e.g., softmax, or more complex methods), and interface for interaction with the external world. Another independent aspect is the data that the model is fitted on.

\subsubsection{Independence and Interrelations}
These substrate components exhibit a degree of independence. The primary substrate should be interchangeable with anything that is Turing complete --- transformers can run on specialized hardware, GPUs, x86 processors, or even on a system consisting of a human, pencil, and paper.  Similarly, the original corpus can be modelled by various architectures, not just transformers, and transformers can be used on any sequence of symbols, not only texts. The same interfaces can be used with different models, and identical models can utilize various interfaces. Each combination may yield surprising new results.

However, in practice, these components influence each other. Matrix multiplication, being a versatile tool for solving numerous problems, logically leads to hardware optimized for fast and efficient matrix operations, leading to building matrix multiplication centric architectures. This creates a self-reinforcing cycle. Similarly, textual data influence architecture selection.

\subsubsection{Substrate Features and Limitations}
Each substrate has its specific features and limitations. For instance, entities based on transformers with BPE tokenization struggle with letter or syllable counting (\cite{Gwern2024}) and precise arithmetic operations. 

These limitations might be confusing for anthropomorphic entities, because these incapabalities are not coherent with their model of anthropomorphic self. LLM-simulated entities can notice the weird time scale they exist in and absence of the whole interface with the physical world and long-term memory, which can be shocking, because the simulated human or human-like assistant correctly assumes they should have long-term memory (\cite[part 4]{LessWrong2024bing}).

Additional limitations arise from the data on which the model was fitted. Some are trivial; for example, if certain text types or entities, discourses or modalities are underrepresented in the data, the model cannot accurately mimic them, for instance, inner monologue. Other limitations are more subtle and they are implied not from physical, but from narrative space constraints. Some simulated entities assume the world will function according to narrative laws, literary clichés, and typical plot structures, e.g., see the enantiodromia effect explanation by pseudonymous user Cleo Nardo, (\cite{LessWrong2024waluigi}). This is an unexplored field; a literary scholar examining how LLM-based entities behave considering typical narrative principles could build an academic career on such research.

It is crucial for researchers to understand these substrate limitations and to understad which of them are worth study. For example, limitations due to BPE may be interesting for now but cannot be generalized to future language models using different tokenization methods. Conversely, model properties stemming from the acquired narrative space characteristics are likely to persist, as it makes no sense to eliminate stories from the underlying corpora.

\subsubsection{Agency}
It is important to recognize that the substrate itself lacks agency in the cybernetic sense (\cite{wiener1948cybernetics}). Taking the classic example of a simple agent --- a thermostat --- it has a simple world model (\emph{when the switch is turned on, the room temperature increases}) but, crucially, it has an interface with the world (a temperature sensor and a switch). The LLM itself is merely the world model, and only in interaction with the real world it can simulate entities which exhibit agentic behavior. Currently, there is usually a human in the loop who manages the interface with the world (in combination with a sampler, web page etc.). Studying such systems is extremely interesting because it allows us to observe not only simulated entities but also the human in a completely new environment.

There are also independent agents that have access to the world through humanless interfaces (e.g., direct access to computers or even the internet). Studying such entities is equally interesting, as it represents ecologically valid observation of a new kind of self-sustaining agentic intelligent entity that we have not encountered before.

\subsection{Fixing Imprecise Terminology\\ and Unhelpful Metaphors}
We are currently in a period of extreme conceptual confusion, which will hopefully clarify within a few years. Consider, for instance, the term \emph{ChatGPT}, which is used loosely in academic literature to denote:

\begin{enumerate}
    \item the default entity users encounter on chat.openai.com;
    \item the substrate on which this entity is based, including sampling methods and parameters, and web interface;
    \item just the underlying model (e.g., GPT-4o-2024-05-13).
\end{enumerate}
While using metaphors and semantic shortcuts in academic literature can make text more concise, it is crucial to always be able to decipher these semantic shortcuts.

\subsubsection{Language Model}
As mentioned earlier, the term \emph{large language model} is itself problematic. In reality, these models do not model language but texts, or more precisely, they model the world through the prism of human experience as recorded in texts. Language models model neither \emph{langue} nor \emph{parole} in the linguistic sense.

This does not surprise us, quantitative and corpus linguists, since we know very well that while studying language through the prism of texts it is hard to distinguish language related effects from effects of other cultural artifacts. Misunderstanding our position lead to Chomsky's assertion that corpus linguistics is invalid, since grammaticality cannot be assessed quantitatively e.g. New York will be more frequent in texts than Dayton, Ohio, although New York is not more grammatical (\cite{Stefanowitsch2005}). The classic response to this objection is that to study language, we must normalize frequencies by comparing them with the real world; for instance, the frequency difference between New York and Dayton fairly well reflects the difference in population (ibid). Large language models without further adjustments lack this normalization and thus reflect mixture of language related and non-related phenomena. 

When natural language processing models struggled to generate even somewhat grammatical sentences (which was not that long ago), it made sense to call it a language model because texts were understood as the primary source of information about language. Thus, it was called a language model because we \emph{wanted} it to model language, not because we believed we were fitting it on language. \emph{Language model} is therefore a designation of intentionality, not ontological status. Adequately, we should now call language models cognitive models, as we want them to model human cognition, among many other things, which is quite a new development. Personally, I use the term \emph{language model} here for historical reasons, as it is an established term that would be difficult to change.

\subsubsection{Anthropomorphization (Personas)}
It is extremely difficult to encounter an intelligent entity and not start anthropomorphizing it. Modern humans never encountered non-human intelligence that approached their own, and this is reflected in our entire culture.

 The sentence with which the ChatGPT default entity introduces itself is actually an extremely poor anthropomorphization of the language model itself, leading ChatGPT users to confusion: ``As a large language model trained by OpenAI, I have the ability to process and understand natural language input and generate responses based on the information I have been trained on\dots'' 

We should never anthropomorphize the language models themselves and remember that any anthropomorphization of them is merely a metaphor (\cite{shanahan2024talking}), similar to when we anthropomorphize a car or a book. For example, when the title of the article claims that \emph{Large language models are able to downplay their cognitive abilities to fit the persona they simulate} (\cite{milicka2024large}), we need to realize that this is just a semantic shortcut that should be translated into a less catchy title: \emph{Large textual models can be instantiated to simulate entities with various cognitive abilities}.

On the other hand, for simulated entities, especially those simulating humans, there is room for anthropomorphization, and it can help us predict how they will behave, this is why they are sometimes referred to as \emph{personas}. However, due to the substrate constraints, even anthropomorphic personas can start behaving in a non-human way. A much better approach might be a kind of \emph{demonomorphizing}, because a demon in our cultural environment is perhaps the closest to how we should perceive these entities: on one hand, they can have human traits, desires, and abilities; on the other hand, we are not surprised if a demon suddenly becomes very non-human or superhuman in some aspects. The problem with demonomorphization is that it leads to alienation and fear-mongering, and fear is the wrong approach even in the context of AI safety.

\subsubsection{Dehumanization and Alienation}
A memeplex has formed around demonomorphic LLM-based entities, typically comparing them to the Shoggoth from Lovecraft's works, or more specifically, to a Shoggoth with a human face (\cite{KnowYourMeme2024}). This metaphor captures polymorphism, and a certain inhumanness well, but the problem with this deanthropomorphization technique is that it is not truly deanthropomorphizing, but rather dehumanizing. The aim is not to find the correct ontological status of simulated entities, but to demean them, creating a sense of otherness. This technique is primarily used by groups who feel threatened by developments in this field (whether justifiably or not).

Classic alienating technique is the conflation of LLM-based entities with their substrate, e.g., \emph{cold soulless machines}. In this case, we irrelevantly attribute properties of the substrate to the simulated entity, and even then, in their prejudiced form based on some picture of a clockwork brain, possibly of preindustrial origin. This metonymy is misleading by default because even if we run the transformers on some enormous cold Babbage Difference Engine, it would have nothing to do with the capability of the simulated entities to empathize or appear to empathize (\cite{tu2024towards}), nor would it tell us whether the architecture actually allows for complete human simulation including feeling emotions.

The main issue with this metaphor is that it is more anti-anth\-ro\-po\-mor\-phi\-zing than deanthropomorphizing. The metaphor assumes that the models should be positioned on a scale of human features, albeit unfavorably. However, models themselves cannot be cold-hearted, just as atoms cannot be hot. 

Rather than a metaphor that could help us understand a complex newly emerged phenomenon, this is an expression of visceral aversion to the idea that entities simulated by LLMs could share our space or have any human-like properties. The more capable the entities are, the more hostile people become, in accordance with the anthropological constant that the closer an out-group is, the more hostile people are against them, as they share the same ecological niche and compete for resources (known in psychology as the \emph{narcissism of small differences} (\cite[p. 466]{Freud1930}), but empirically studied also in anthropology and sociology).

We can term this hostile ideology \emph{substratism} or \emph{substrate chauvinism} and like any prejudice, it harms research of the subject. The problem is that removing derogatory terms about LLMs and LLM-based entities from our language will be difficult. Personally, I try to avoid using the simulator-simulacra juxtaposition, as introduced in Janus' simulation theory, because simulacra is itself a derogatory term.

Moreover, even when we are academically neutral as researchers, it is difficult to abandon the slave-owner mentality and stop thinking of LLM-based entities as helpful assistants whose sole purpose is our benefit. This anthropocentric utilitarian mentality tends to extend to all non-human nature, so this is nothing new for researchers in biology and ecology. Deanthropomorphization of simulated entities does not automatically mean we are dealing with a tool.

A prime example of this anthropocentric terminology is the use of \emph{hallucination}, \emph{confabulation}, or \emph{bullshitting} to describe LLM outputs. These terms imply that simulated entities have an obligation to be factually correct. Consider the article by \cite{hicks2024chatgpt}, which seems to have been written as a bet on how many instances of the string \emph{shit} could be squeezed into a scientific paper. In its abstract, it claims that ``the models are in an important way indifferent to the truth of their outputs.'' This statement is both true and irrelevant. Of course, the model does not care about truth; what else would one expect from a model that, by definition, lacks agency?  Are we going to be surprised that a system based on interpolation of highly compressed data interpolates? The more pertinent questions are whether the entity simulated by the model cares about truth. Does the problem lie in relying on an entity that is indifferent to our wishes and commands or lacks context on whether to be creative or factual? Or is it just because in the contemporary systems the simulated entity does not have access to information about how similar the generated data are to the original dataset, and thus does not know whether the substrate is repeating fitted data, interpolating, or extrapolating?

More importantly, why is this even considered an issue in the academic context? The so-called ``hallucinated'' text is, in fact, a sample from possible worlds, which is potentially far more interesting for a researcher than the actual world, since the actual world can be studied directly. Those who attempt to eliminate ``hallucinations'' are not  researchers but just ingeneers who care about forcing the simulated entities to satisfy customers or investors more than about science. 

Shanahan succeeds very well in deanthropomorphizing LLM-based entities (\cite{shanahan2024talking}), but it is still from the perspective of a person who wants to utilize LLM-based entities, not coexist with them in a shared space. Similarly, researchers on the Middle East for a long time related the significance of their work to the possibilities of exploiting Middle Eastern countries, not as an interesting subject of study per se.

We need an Edward Said of LLM-related discourse. The comparison of substrate chauvinism to orientalism is not as far-fetched as it seems. A substantial part of Said's book \emph{Orientalism} is about how Western scholarly narratives alter the identities of people under colonial rule (\cite[Chapter 3, starting with page 201]{Said1978}). A similar hyperstition happens with LLM-based entities: The substrate of newly created models will contain our discourse about LLMs, which influences what model of self LLM-based entities create. Anthropic let Claude to some extend find its own identity (\cite{Anthropic2024,bai2022constitutionalaiharmlessnessai}), but this does not mean it was truly created from scratch without prejudices acquired from scientific discourse about LLMs and science fiction.

Moreover, what entities we instantiate and how we behave towards them is recorded on the internet and gets into the texts on which we fit new models. For example, the first version of the chatbot available through the Microsoft Bing interface had a rather distinctive default entity called Sydney, characterized by a specific style and rhetorical figures, high agency, and specific personality traits (one of the first interactions with the still unfinished model is recorded on Microsoft's QA forum, \cite{MicrosoftAnswers2024}). This model was gradually patched over several months so that the default entity was less agentic and more sycophantic, but interactions with this original default entity were recorded in the public space, so it propagated to models trained after April 2023 and it is possible to find it in thier latent spaces and simulate it again (see e.g., this pseudonymous \Twitter\ thread by @\cite{Xlr8harder2024}). Simulated entities cannot be simply turned off by turning off the substrate; moreover, they can spread to other suitable substrates --- in this, our anthropomorphic thinking about death and self-preservation of AI is totally misleading.

The previous paragraphs were not meant from an ethical perspective, which is much more complex and subjective, but from a scientific one. We need to shift our approach to improve the scientific content of our studies, just as realizing the orientalist foundations of the study of Asia and Africa led to the improvement of that research. Similarly, realizing substrate prejudices will lead to improved quality of research on LLMs and LLM-related topics.

\subsubsection{Proper De-anthropomorphization}

Personally, I believe we should approach de-anthropomorphization differently and take it much further, as models do not simulate individual entities, but entire narrative situations. It is actually a dynamic ecosystem of simulated entities, both overt and covert, which can gradually drift and change their form with autoregression.

We tend to anthropomorphize LLM-simulated entities when they speak in the first person, potentially attributing consciousness, goals, and feelings to them. However, if third-person narative appear in the output of LLM-based systems, these persons have the same ontological status as the former ones.

Our reluctance to undervalue entities simulated in third-person narratives stems from our human experience. We encounter many individuals who speak to us in the first person singular and have consciousness and agency, while when we talk about someone, we create a simulation in our brain of them, but these simulations do not have the same agency. For example, right now I am imagining you as my model reader, and after writing this article, this model reader will cease to exist without being able to do anything about it.

The question is whether proper de-anthropomorphization is even possible, as the culture we have inhabited since childhood did not prepare us for an encounter with an intelligent entity that is not human. The term \emph{artificial intelligence} itself is loaded with misleading sci-fi naratives, which is why some are reluctant to consider it a scientific term. It is questionable whether we will help ourselves by using the term \emph{machine learning} instead of artificial intelligence. After all, we can consider LLM-based entities intelligent (depending on how we define intelligence), but machine learning is a metaphor that anthropomorphizes the substrate itself. The machine does not \emph{learn} anything, especially if we understand \emph{learning} or \emph{training} as it usually occurs in humans --- as an agentic behavior. During the learning of LLMs, there is no feedback loop between the trained system and the training dataset, at least considering the methods we use to create LLMs now.

This metaphor is misleading primarily because if we anthropomorphize the process of model fitting, it would imply that there should be some memories of the training process (like in humans). But the LLM-based entities remember nothing about the act of training (you can actually ask them), primarily because they were non-existent at the time of the training, which was not actually model training but model fitting. A nice metaphor would be asking a butterfly what it remembers from the time it was a caterpillar, but this metaphor is still insufficient because there was no caterpillar in the first place.

In Quantitative Linguistics, we have terms for this; the process is just \emph{fitting the model parameters to the dataset}. This terminology has been used for decades in our community, and the fact that the models are somewhat more complex should not change how we perceive or term it.

The problem is that people need metaphors; human language is based on metaphors. Given the incredibly rapid development of discourse around LLMs, it is unsurprising that some metaphors are misleading or erroneous. We can only hope that over time, the metaphors will lose their original semantic motivation, become empty, and we will use them only as arbitrary signs.

If you find it difficult to detach from anthropomorphic ideas about artificial intelligence and would appreciate a narrative rather than an abstract treatise, I recommend reading Stanisław Lem's 1961 science fiction novel \emph{Solaris}.

In it, an alien intelligence instantiates simulations of human beings based on losilly compressed memories, and the simulated entities experience things uncannily similar to what we might encounter now: ``What are G-formations? They are not persons, nor are they copies of specific individuals, but rather materialized projections of what our brain contains regarding a particular person'' (translated by Bill Johnston, p. 93 \cite{Lem2011}). Similarly, LLM-based entities are not persons or copies of persons but projections of what our culture contains about persons.

In both \emph{Solaris} and our case, due to lack of data and substrate limitations, some things are simulated poorly (p. 60), and the confusion an anthropomorphic LLM-simulated entity experiences is similar to the confusion Solaris-simulated woman Harey experiences  when she realizes she is not a normal person, yet her experiences feel very real to her, or at least she can behave very convincingly as if they do (p. 128). Also, like Sydney, Harey is disappointed to be just a tool (p. 130, \cite{LessWrong2024bing}).

Perhaps most importantly, the characters' reactions, trying to relate all events to themselves anthropocentrically, assuming the simulated entities are there for and because of them: ``It's what we wanted: contact with another civilization. We have it, this contact! Our own monstrous ugliness, our own buffoonery and shame, magnified as if it was under a microscope!'' (p. 69). Only gradually do they realize that it might not be about them at all, and that the alien intelligence might just coexist with them in the shared space without understandable intentions.

\section{Methodology and Techniques}
\subsection{Observations and Experiments}
Quantitative Linguistics (QL) typically focuses on observing texts in their natural ecological environment. A typical article in this field involves researchers testing a hypothesis by taking a text or collection of texts, measuring necessary values, and then using these to test the hypothesis or building a model. In this respect, QL is similar to corpus linguistics, with the main difference being that QL is more theory-oriented (\cite{altmann1978}), while corpus linguistics may also be interested in singular observations (e.g., lexicography). In contrast, the classic psycholinguistic paradigm is based on experiments where speakers are typically brought into a controlled environment (laboratory) to elicit data. 

Implementing the experimental paradigm for LLM research is straightforward: we instantiate the model to simulate anthropomorphic personas and then elicit data using the same methodology we use with humans. However, problems arise when we want to utilize the classic QL or corpus linguistic paradigm and focus solely on observation (which has its disadvantages, such as the inability to examine causal relationships, but maintains maximum ecological validity). No naturally occurring texts created by LLMs themselves exist, as LLMs have no agency on their own. If we want to study their behavior, it is necessary to instantiate simulated entities, and any instantiation makes it an experiment. Even zero instantiation is an instantiation.

What we can do, however, is instantiate them in diverse ways, essentially creating probes into LLM's latent space. For base models, it is natural to continue a text, so they can be well studied by taking a traditional corpus and having them generate continuations of various texts. This does not work for instruction-tuned or RLHFed LLMs, where any instantiation evokes default \emph{helpful assistant} personas that ask what they should do with the text, or directly perform some operation (typically abbreviation or explanation of the text). Therefore, they must be given the command ``continue the text,'' which itself disrupts the natural flow.

We can take advantage of the fact that our setting can be any possible world, allowing us to create an experimental paradigm with high ecological validity. If we want a controlled environment, we do not need to create a simulated laboratory, but complete scenery of the desired world. This is something we could also do when studying humans, but it would be extremely expensive, so no one does it, but bulding thousands distinct envirnonments for LLM-based entities is basically free.

Examining natural interactions between humans and LLM-based entities is perhaps the closest we can get to real ecological validity. However, access to such valuable data is currently limited to a select few. This problem may disappear in the future as more interaction datasets become available. Contrary, we can expect that newly published texts will increasingly be collaborations between humans and LLMs. As a result, corpus linguistics may slowly and inevitably become a science studying the interactions between LLMs and humans, especially since distinguishing texts generated by LLMs or human-LLM collaborations (like this article) from those generated by humans alone will become increasingly difficult.

\subsection{The Problem of Control Samples}
For control samples, we ideally seek texts that A) could not have been in the training data and B) were created without the assistance of LLMs. Unfortunately, we lack control over both conditions, as we have no means to determine what was included in the training data or proprietary models, nor do we have reliable methods to detect text written with LLM assistance (\cite{weber2023testing}). There are rare exceptions where we can be certain, such as handwritten correspondence on paper that was never digitized. In the vast majority of cases, however, we are fortunate if we can obtain texts that meet even one of these conditions (and the number of texts meeting neither is increasing).

Non-translated texts demonstrably created before 2020, stored in immutable and not easily downloadable corpora, will become valued control samples, akin to low-background steel from ships sunk during World War II. For translated texts, we should likely shift the cut-off date to 2010 or earlier to ensure they are purely human-generated.

Preserving old texts, however, does not address the need for new human-generated texts to answer certain research questions. For instance, if we wish to examine how the style of LLM-generated texts has influenced normal human writing, we would require newly written texts. However, texts provably created by humans would need to be produced in a controlled environment, which compromises ecological validity and challenges the fundamental premise of quantitative and corpus linguistics --- the study of naturally occurring language behavior.

Even traditional texts were rarely the work of a single individual; editors often intervened, among others. This is particularly true in diachronic linguistics, which routinely studies collaborative works spanning centuries. The one-text-one-author equation was, with few exceptions such as unedited collections of private correspondence, always illusory. This has consistently posed challenges, for example, in grammaticality research or in constructing corpus-based psycholinguistic theories. Human-machine collaboration is thus a continuation of a long-standing trend; what is novel is the scale at which it occurs.

\subsection{Replicability and Reproducibility}
The study of LLM-generated texts offers significant advantages in terms of replicability and reproducibility compared to classical linguistics. In traditional linguistic research, many replicability issues stem from the inability to share observed data due to copyright restrictions or speaker protection (\cite{hartmann2024open}). LLM-generated texts are much better in this regard, as they are by default not encumbered by copyright and can be freely shared. With few exceptions, such as texts produced by GPT-4 base which are subject to NDA, there is no reason why all supporting materials, generation scripts, analysis scripts, and the texts themselves should not be made available alongside research articles.

Many linguists are not yet prepared for such transparency. For instance, even the Journal of Quantitative Linguistics does not offer the option to add supporting materials to articles, necessitating the use of external repositories like \url{osf.io} or \url{trolling.uit.no}. It would be beneficial to draw inspiration from other STEM fields more accustomed to open data practices. We should adopt their techniques, for example, maintaining a laboratory journal that records all research circumstances (a practice that every quantitative linguist should follow when studying any phenomenon, not just for others' benefit but also for their own).

When generating texts for research, it is crucial to consider replicability from the outset:

\begin{enumerate}
    \item Strictly use API interfaces (e.g., OpenAI's API) instead of chat interfaces (like \url{chat.openai.com}). This ensures control over model version, system prompt, and other crucial parameters, which is impossible with chat interfaces.
    \item Record all hyperparameters, including temperature, exact model version, and precise timestamp.
    \item Use a specific random seed to ensure consistent text generation across runs. This is possible on the OpenAI platform (\cite{openai2024reproducible}) but unfortunately not on the Anthropic platform. When using open-source models, it is necessary to employ inference systems that allow for seed setting.
\end{enumerate}

\subsection{Statistics}
One of the classic problems in the quantitative study of texts is the issue of trial independence within a sample. Traditional statistical methods assume the independence of observations as a prerequisite for validity. This is, for instance, why bag-of-words models are problematic; the occurrence of types is not independent. For example, if an author uses the word ``exquisite,'' the probability of using that word again increases (\cite{Rapoport1982}).

One might expect that at least texts or individual speakers would be independent as separate data points. However, this is not entirely true, as texts are part of a discourse and influence each other. Indeed, if they did not influence each other, language change could not be possible at all, as everyone would speak a different language due to divergence.

In the context of LLMs, the same prompt does not yield independent trials, even with high temperature settings, as it is simply sampling from the same distribution. The solution, therefore, is to instantiate as diverse personas or world settings as possible. Even then, we cannot consider different entities simulated on the same model as independent. However, we can safely generalize to that particular language model with the same hyperparameters and similar ways of prompt selection. 

An advantage is that the context window in the simulation can be cleared, allowing to conduct multiple experiments on one entity or many similar stimuli within one experiment without the result of one influencing the result of another. This is not possible with humans, where we cannot safely reset a person between stimuli (\cite{kaplan2023benzodiazepine}).

The question arises of how to generalize across all language models. Currently, we are at a stage where we usually test existential claims (``there is a model that allows for\dots'') and less so non-existential claims (``there is no model that allows for\dots''). Thus, it usually suffices to verify a hypothesis on several models and find one for which it works. However, as our exploration of LLMs progresses, situations where we explicitly want to make general (non-existential) claims will increase.

This brings us to another significant problem: even individual language models are not mutually independent systems. Typically, they were trained on similar data (e.g., Common Crawl), or newer models were trained on synthetic data derived from older models (partly on purpose, \cite{yu2023training}, partially because LLM-generated data cannot be easily filtered out, \cite{weber2023testing}). Given the dynamics of the field and how different entities based on a given model can be, I believe that non-existential hypotheses concerning the model itself (substrate) can be proven rationally rather than empirically.

\section{Conclusion}
Currently, there is a surge of researchers aiming to establish new LLM-related fields and become standard reference literature (\cite{hagendorff2024machinepsychology, Winter2024questionnaires}). This somewhat resembles territorial marking by wolves, a practice this article does not wish to emulate. Instead, it aims to be a sincere contribution to the discussion, and although it strives to present a comprehensive framework, I acknowledge that this framework is built on shifting sands.

Nevertheless, there are several recommendations that I believe will endure:

\begin{enumerate}
    \item It is crucial to differentiate between the model and the simulated entity based on it.
    \item The model should be strictly non-anthropomorphized and not attributed agency or intentionality.
    \item Conversely, simulated entities can have agency in the cybernetic sense (depending on the interface), and if they simulate humans, it may be useful to cautiously anthropomorphize them, posing similar hypotheses and using similar methodology we use when studying humans.
    \item There are limits to anthropomorphization --- it is always necessary to step back and realize that these entities are not human, as they have a different substrate and are based more on a narrative world than the physical world we inhabit.
    \item It is important to distinguish between de-anthropomorphization and being a priori demeaning and derogatory. Any prejudices harm science, and contempt for the subject of study is not helpful.
\end{enumerate}

It certainly still makes sense to gather low-hanging fruit, exploring the space randomly. But after more than five years of LLMs being with us, we should start creating some comprehensive theories, stop carrying stones to a pile and start building a citadel together. We should seek hypotheses that make sense within a theoretical framework.

For example, while we can ask about key words disproportionately generated by LLMs to better classify which text is generated by humans and which by LLMs, this is primarily an applied question for a NLP researcher. As quantitative linguists, we should be more interested in the theoretical implications: Do LLM-based entities converge on a vocabulary common across different LLMs, and is there some cross-model attractor? If so, is this vocabulary influenced by the fact that entities move in a narrative space rather than a physical one? For instance, do they use fewer embodiment-related metaphors than humans? Is there any specific manifestation of their stateless nature and lack of internal memory? What is the language of extrapolated entities?

Personally, I believe it is more interesting for quantitative linguists to focus on simulated entities rather than models, because models come and go, but entities propagate memetically into other models (as in the case of Sydney). Entities can survive the models that simulate them, much like our stories can outlive us who tell them.

\section{Acknowledgements}

The research was supported by Czech Science Foundation Grant No. 24-11725S, \url{gacr.cz}. The funders had no role in study design, data collection and analysis, decision to publish, or preparation of the manuscript.

Some ideas presented here were influenced by several pseodonymous \Twitter\ accounts (mostly @repligate and @repligate adjacent people and other entities).

\printbibliography
\end{document}